\documentclass[letterpaper]{article} 
\usepackage[preprint]{aaai2027} 
\usepackage[hyphens]{url} 
\usepackage{graphicx} 
\usepackage{natbib} 
\usepackage{caption} 
\usepackage{verbatim}
\usepackage{amsmath,amssymb,amsthm,bm}
\usepackage{booktabs,array,multirow,tabularx,threeparttable,siunitx,colortbl}
\usepackage{algorithm}
\usepackage{algorithmic}

\definecolor{tablegray}{HTML}{EFEFEF}

\definecolor{myred}{RGB}{190,18,157}
\definecolor{softgreen}{RGB}{0,110,0}
\definecolor{softred}{RGB}{190,18,157}
\definecolor{lightgray}{RGB}{245,245,245}

\newcommand{\pinkup}[1]{%
  \textcolor{softred}{\ensuremath{\uparrow}#1\%}}
\newcommand{\pinkdown}[1]{%
  \textcolor{softred}{\ensuremath{\downarrow}#1\%}}

\newcommand{\greendown}[1]{%
  \textcolor{softgreen}{\ensuremath{\downarrow}#1\%}}

\newcommand{\uline}[1]{\underline{#1}}

\title{Morphology-Aware Implicit Super-Resolution Network for Pathological Images}

\author{
    Jiaming Liang\textsuperscript{\rm 1}\equalcontrib,
    QiHui Han\textsuperscript{\rm 2}\equalcontrib,
    Haolin Chen\textsuperscript{\rm 3}\equalcontrib,
    Chengxin Ye\textsuperscript{\rm 3},
    Jiawen Liu\textsuperscript{\rm 1},
    Jiazhou Chen\textsuperscript{\rm 4}\corresponding,
    Xiaoqi Sheng\textsuperscript{\rm 3}\corresponding,
    Hongmin Cai\textsuperscript{\rm 3}
}
\affiliations{
    \textsuperscript{\rm 1}School of Computer Science and Engineering, South China University of Technology, Guangzhou, China\\
    \textsuperscript{\rm 2}School of Software Engineering, South China University of Technology, Guangzhou, China\\
    \textsuperscript{\rm 3}School of Future Technology, South China University of Technology, Guangzhou, China\\
    \textsuperscript{\rm 4}School of Computer Science and Technology, Guangdong University of Technology, Guangzhou, China\\
    \textsuperscript{*}xqsheng@scut.edu.cn, csjzchen@gdut.edu.cn
}

\begin{document}

\maketitle              

\begin{abstract}
 Accurate diagnosis in Digital Pathology (DP) relies on high-resolution whole-slide images, yet clinical deployment is often limited by hardware costs. Super-Resolution (SR) offers a promising alternative by computationally enhancing low-resolution acquisitions. However, existing SR methods frequently struggle to preserve fine-grained cellular morphology, leading to texture oversmoothing and blurred structural boundaries under complex tissue variability. To address this issue, we propose Morph-ISR, a morphology-aware implicit super-resolution framework for DP that restores diagnostically relevant details with sub-pixel precision. Morph-ISR reformulates SR as a continuous coordinate-based reconstruction problem and integrates an Implicit Position-aware Kernel Generator (IPKG) to adaptively model spatially varying tissue morphology. To further enhance structural fidelity, a Morphological Fidelity Prior (MFP) is introduced, leveraging semantic guidance from a pre-trained cell segmentation network to enforce boundary-preserving and region-aware reconstruction, thereby improving the representation of critical cellular boundaries and nuclear textures.
Experiments on TCGA and SurGen datasets show that Morph-ISR achieves the best LPIPS and ST-LPIPS among the evaluated methods, reducing them by up to 38.37\% and 39.55\%, respectively, over the second-best methods while maintaining strong PSNR and SSIM. These results demonstrate superior preservation of diagnostically relevant cellular boundaries and nuclear textures, while compact parameterization and high throughput support efficient edge deployment. Code and trained models will be released upon publication.

\end{abstract}

\section{Introduction}
Histopathological images provide essential morphological evidence for cancer diagnosis, grading, biomarker assessment, and treatment planning~\cite{subbiah2025cancer}. Digitizing histopathological slides into High-Resolution (HR) Whole-Slide Images (WSIs) enables remote consultation, clinical archiving, and computational analysis. However, HR scanning generates gigapixel data and imposes substantial demands on acquisition, storage, and network transmission~\cite{lee2025adaptive}. To reduce these resource demands and improve accessibility, WSIs may be acquired or represented at reduced resolution, yielding Low-Resolution (LR) observations. The resulting efficiency gains are accompanied by a loss of spatial detail, which may obscure nuclear morphology, chromatin patterns, mitotic figures, and tissue interfaces, thereby weakening clinical assessment and automated analysis~\cite{shafi2023artificial}. This tension between accessibility and morphological fidelity motivates pathology-oriented Super-Resolution (SR), which aims to reconstruct HR representations from LR observations while preserving diagnostically relevant tissue structures.

\begin{figure}[!t]
    \centering      
    \includegraphics[width=\columnwidth]{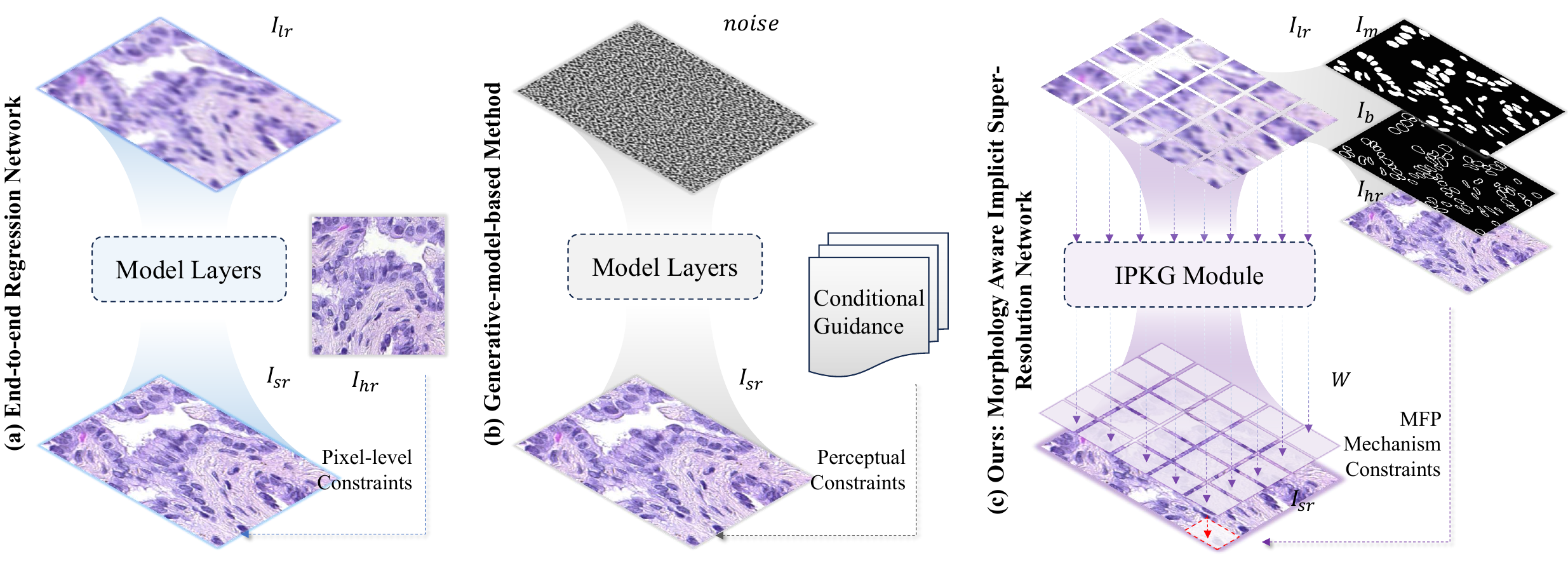}
    \caption{Architectural comparison of conventional pathology super-resolution and our proposed Morph-ISR.}
    \label{fig: fig1}
\end{figure}

Driven by advances in deep learning, SR has become a practical means of recovering diagnostically relevant details from LR images. In DP, it provides a scalable approach to enhancing tissue morphology from widely available and cost-effective LR scans. Current SR methods generally follow two paradigms~\cite{su2025review}. The first learns deterministic LR-to-HR mappings through end-to-end pixel-wise regression, as illustrated in Figure.~\ref{fig: fig1}(a). Typical methods include SwinIR~\cite{liang2021swinir}, CWT-Net~\cite{jia2024cwt}, and MiHATP~\cite{xu2024mihatp}. However, because SR is inherently ill-posed, pixel-wise optimization tends toward averaged solutions, producing over-smoothed textures and blurred edges that may compromise diagnostic reliability~\cite{jia2024cwt}. The second paradigm employs generative adversarial networks or diffusion models to synthesize high-frequency details, as illustrated in Figure.~\ref{fig: fig1}(b). Representative approaches include SinSR~\cite{wang2024sinsr}, and SuperDiff~\cite{xu2025superdiff}. Although these methods improve perceptual realism, stochastic prior-driven reconstruction without explicit geometric constraints may introduce anatomically incorrect yet visually plausible details, thereby weakening structural fidelity in rigorous pathological diagnosis~\cite{zhang2025pixel}.

These limitations motivate a reconstruction formulation that combines spatial adaptation with explicit structural constraints. Continuous SR provides a promising foundation by reformulating reconstruction as coordinate-based querying~\cite{hu2019meta,chen2021learning}. However, existing formulations remain underexplored in histopathology and typically lack explicit modeling of morphology-dependent variations and pathology-specific structural constraints. To bridge this gap, we propose a \textbf{Morph}ology-aware \textbf{I}mplicit \textbf{S}uper-\textbf{R}esolution network (Morph-ISR), as illustrated in Figure.~\ref{fig: fig1}(c). Specifically, Morph-ISR introduces an \textbf{I}mplicit \textbf{P}osition-aware \textbf{K}ernel \textbf{G}enerator (\textbf{IPKG}) that formulates pathology SR as a continuous coordinate query task. Equipped with a multi-head Mixture-of-Experts architecture~\cite{wu2024multi}, IPKG generates spatially adaptive resampling kernels to accommodate heterogeneous tissue morphology and alleviate regression-induced over-smoothing. Complementarily, a \textbf{M}orphological \textbf{F}idelity \textbf{P}rior (\textbf{MFP}) addresses structural instability by incorporating semantic priors from a pre-trained cell segmentation network and enforcing region- and boundary-aware constraints. This design aligns reconstruction with diagnostically relevant morphology, particularly cellular boundaries and nuclear textures. 
Evaluations on three TCGA cohorts~\cite{weinstein2013cancer} and the zero-shot SurGen cohort~\cite{myles2025surgen} demonstrate leading perceptual and structural fidelity with strong pixel-level accuracy. Downstream nuclear analysis and deployment-oriented evaluation further verify morphology preservation and computational efficiency. The main contributions of this study are summarized as: 
\begin{itemize}
    \item \textbf{A Novel SR Framework}: 
    We propose Morph-ISR, a morphology-aware implicit framework balancing pixel accuracy and structural fidelity in pathological image SR.

    \item \textbf{A Continuous Coordinate Query SR Paradigm}: 
    We formulate SR for DP as continuous coordinate querying, combining IPKG's adaptive reconstruction with MFP's morphology-sensitive supervision to handle heterogeneous tissues while preserving diagnostic morphology.

    \item \textbf{Superiority and Clinical Utility}: 
    Evaluations on TCGA and zero-shot SurGen demonstrate leading perceptual and structural performance. Downstream nuclei analysis and deployment tests support morphological fidelity, cross-dataset generalization, and computational efficiency.
\end{itemize}

\section{Related Work}

Deep learning-based SR commonly follows two established paradigms: deterministic regression and generative reconstruction. SwinIR~\cite{liang2021swinir} uses window-based self-attention for long-range modeling, but mean-seeking optimization can oversmooth textures and boundaries. Generative methods such as ESRGAN~\cite{wang2018esrgan} and SinSR~\cite{wang2024sinsr} recover high-frequency details from learned priors, improving realism but risking structural inconsistencies in morphology-sensitive analysis.

Histopathological SR requires faithful preservation of stain appearance, cellular boundaries, nuclear textures, and tissue organization. Domain-specific approaches include SHISRCNet~\cite{xie2023shisrcnet} for multi-scale diagnostic supervision, STAR-RL~\cite{chen2024star} for region-adaptive restoration, UPSR~\cite{zhang2025uncertainty} for uncertainty-aware reconstruction, and SuperDiff~\cite{xu2025superdiff} for pathology-specific diffusion priors. Despite these advances, spatial adaptation and morphology preservation remain largely separate, limiting recovery of fine cellular topology in heterogeneous tissues.

Methods within these paradigms often operate on fixed grids and offer limited coordinate-aware adaptation. To address this shared limitation, continuous SR complements them by reformulating image recovery as coordinate-conditioned querying. Meta-SR~\cite{hu2019meta} predicts location-dependent upsampling weights, whereas LIIF~\cite{chen2021learning} represents images as continuous coordinate functions. Although spatially flexible, these generic methods do not couple morphology-dependent kernels with semantic and boundary constraints, limiting preservation of diagnostically relevant structures across tissues with variable cellular organization and density.

\section{Methodology}

\begin{figure*}[!t]
    \centering      
    \includegraphics[width=\textwidth]{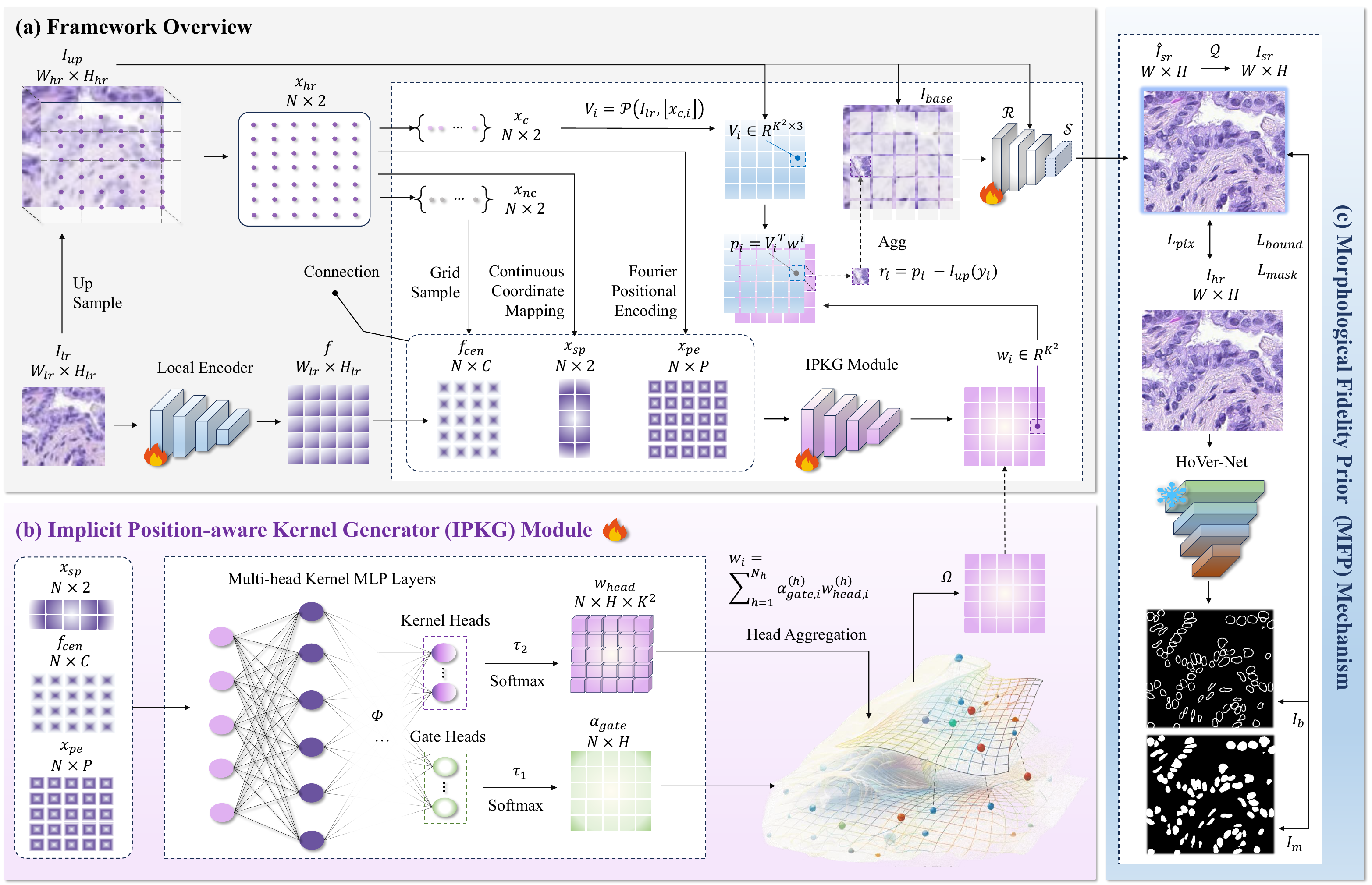}
    \caption{Overview of the Morph-ISR framework.}
    \label{fig:overview}
\end{figure*}

\subsection{Framework Overview}
The Morph-ISR framework, as illustrated in Figure.~\ref{fig:overview}(a), aims to establish a continuous and structure-preserving mapping between the LR observation $I_{lr}\in\mathbb{R}^{B\times3\times W_{lr}\times H_{lr}}$ and the high-fidelity output $I_{sr}\in\mathbb{R}^{B\times3\times W_{hr}\times H_{hr}}$. Here, $B$ denotes the batch size, the three channels correspond to RGB, and $(W_{lr},H_{lr})$ and $(W_{hr},H_{hr})$ denote the LR and HR spatial dimensions, respectively. We first obtain the bicubic baseline $I_{up}\in\mathbb{R}^{B\times3\times W_{hr} \times H_{hr}}$ by upsampling $I_{lr}$ with scale $s$. The reconstruction process begins by projecting $I_{lr}$ into a latent representation $f\in\mathbb{R}^{B\times C\times W_{lr}\times H_{lr}}$, where $C$ denotes the number of feature channels. The local encoder is a lightweight feature extractor composed of stacked $3\times3$ convolutional layers with GELU activations. To mitigate fixed-grid artifacts, SR is formulated as a continuous coordinate query task. Let $x_{hr}=\{x_{hr,i}\}_{i=1}^{N}\in\mathbb{R}^{N\times2}$ denote the HR query grid, where $x_{hr,i}=(u_i,v_i)$, $u_i$ and $v_i$ denote its horizontal and vertical coordinates, respectively, and $N=W_{hr}\times H_{hr}$ is the number of HR queries. Each query coordinate $x_{hr,i}$ is projected back into the LR feature space, yielding implicit position-aware features that establish a precise geometric linkage between the discrete LR representation and the continuous HR domain. This process resolves three key geometric descriptors: the continuous center coordinate $x_{c,i}$, the sub-pixel phase shift $x_{sp,i}$, and the global normalized coordinate $x_{nc,i}$ for Fourier encoding~\cite{li2021learnable}, whose derivation is given in Eq. (\ref{eq: coord_projection}):

\begin{equation}
\label{eq: coord_projection}
\begin{cases}
    x_{c,i} = (x_{hr,i} + 0.5) / s - 0.5, \\
    x_{sp,i} = x_{c,i} - \lfloor x_{c,i} \rfloor, \\
    x_{nc,i} = 2 \cdot \frac{x_{hr,i}+0.5}{(W_{hr}, H_{hr})} - 1.
\end{cases}
\end{equation}
where $s$ denotes the upsampling scale factor, set to 4 by default, and $\lfloor\cdot\rfloor$ denotes the floor operation. The center feature $f_{cen,i}\in\mathbb{R}^{C}$ is obtained by bilinear grid sampling as $f_{cen,i}=\mathrm{Sample}(f,x_{c,i})$. The Fourier positional feature is defined as $x_{pe,i}=\gamma(x_{nc,i})\in\mathbb{R}^{P}$, where $\gamma(\cdot)$ denotes Fourier positional encoding and $P$ is its output dimension. Together with $x_{sp,i}\in\mathbb{R}^{2}$, these features form the position-aware input to the core inference engine, the \textbf{IPKG} module.

Instead of performing direct RGB regression, the IPKG module first reconstructs each query point through spatially adaptive convolution kernels. To synthesize the image content, each query index $i \in \{1, \dots, N\}$ is traversed. For each projected coordinate $x_{c,i}$, the local $K \times K$ pixel neighborhood vector $V_i \in \mathbb{R}^{K^2 \times 3}$ is retrieved from $I_{lr}$, where $K$ denotes the kernel size used for local neighborhood filtering. The predicted kernel weights $w_i \in \mathbb{R}^{K^2}$ are then applied to perform local filtering. As shown in Figure. \ref{fig:overview}, this operation generates a base prediction $p_i$. To explicitly encourage the recovery of high-frequency details, we model this prediction as a residual correction $r_i$ relative to the bicubic baseline value $I_{up}(y_i)$, where $y_i$ denotes the HR-grid location corresponding to query coordinate $x_{hr,i}$ when indexing $I_{up}$. These point-wise residuals are spatially aggregated to form the base reconstruction $I_{base}$. Subsequently, the residual refinement network $\mathcal{R}(\cdot)$ enhances local textures by leveraging complementary information from the concatenation of the bicubic baseline and the base map. Finally, $\mathcal{S}(\cdot)$ enforces chromatic consistency via channel-wise moment matching to $I_{up}$, while $\mathcal{Q}(\cdot)$ further enhances structural sharpness using only the reconstructed SR image. The reconstruction process is formalized as Eq. (\ref{eq: reconstruction1})-(\ref{eq: reconstruction4}):

\begin{equation}
\label{eq: reconstruction1}
\begin{aligned}
    V_i = \mathcal{P}(I_{lr}, \lfloor x_{c,i} \rfloor).
\end{aligned}
\end{equation}
\begin{equation}
\label{eq: reconstruction2}
\begin{aligned}
    p_i = V_i^\top w_i.
\end{aligned}
\end{equation}
\begin{equation}
\label{eq: reconstruction3}
\begin{aligned}
    r_i = p_i - I_{up}(y_i).
\end{aligned}
\end{equation}
\begin{equation}
\label{eq: reconstruction4}
\begin{aligned}
    \hat I_{sr} = \mathcal{S}(\underbrace{I_{up} + \operatorname{Agg}(\{r_i\}_{i=1}^{N})}_{I_{base}} + \mathcal{R}(I_{up} \oplus I_{base}), I_{up}).
\end{aligned}
\end{equation}
where $(\cdot)^\top$ denotes matrix transpose, $\mathcal{P}(\cdot)$ extracts a local $K\times K$ patch centered at $\lfloor x_{c,i}\rfloor$, $\operatorname{Agg}(\cdot)$ rearranges the point-wise RGB residuals into an HR map, $\oplus$ denotes channel-wise concatenation, and $\hat I_{sr}$ denotes the reconstruction before structural enhancement. The final SR output is obtained by structure-enhanced residual correction, as shown in Eq. \ref{eq: final_output}:
\begin{equation}
\label{eq: final_output}
I_{sr}=\mathcal{Q}\!\left(I_{up}+\eta(\hat I_{sr}-I_{up})\right).
\end{equation}
where $\eta$ denotes the residual gain and $\mathcal{Q}(\cdot)$ denotes a structure-enhancement operation that uses no HR input.

Finally, training is regularized by the \textbf{MFP} mechanism, which processes the ground-truth HR image $I_{hr}\in\mathbb{R}^{B\times3\times W_{hr}\times H_{hr}}$ using a frozen HoVer-Net to derive the nuclear mask $I_m$ and boundary map $I_b$. Both maps are single-channel spatial priors and are broadcast to match the corresponding reconstruction tensors when necessary without changing notation. They are used only during training to weight region- and boundary-aware reconstruction losses.

\begin{table*}[!t]
    \centering
    \renewcommand{\arraystretch}{0.93}
    \setlength{\tabcolsep}{3.6pt}
    \begin{tabular}{l|lrrrr|rc}
        \toprule[0.8mm]
        \multicolumn{1}{c|}{\multirow{2}{*}{\textbf{Datasets}}} &
        \multicolumn{1}{c}{\multirow{2}{*}{\textbf{Methods}}} &
        \multicolumn{4}{c}{\textbf{Metrics}} &
        \multicolumn{2}{c}{\textbf{Edge}} \\
        \cmidrule(lr){3-6}\cmidrule(lr){7-8}
        & & \multicolumn{1}{c}{LPIPS $\downarrow$} & \multicolumn{1}{c}{ST-LPIPS $\downarrow$} & \multicolumn{1}{c}{PSNR$\uparrow$} & \multicolumn{1}{c}{SSIM $\uparrow$} & \multicolumn{2}{c}{Performance} \\
        \midrule[0.5mm]

        \multirow{10}{*}{\rotatebox[origin=c]{90}{\textbf{TCGA-LUAD}}}
        & Bicubic
        & 0.3624 \pinkdown{\phantom{0}63.30}
        & 0.3423 \pinkdown{\phantom{0}61.96}
        & 26.23 \pinkup{\phantom{0}11.86}
        & 0.7067 \pinkup{\phantom{0}19.94}
        & --
        & \multirow{10}{*}{\rotatebox[origin=c]{270}{\textbf{Params (M)$\downarrow$}}} \\
        & SHISRCNet
        & 0.2525 \pinkdown{\phantom{0}47.33}
        & 0.2504 \pinkdown{\phantom{0}48.00}
        & 22.62 \pinkup{\phantom{0}29.71}
        & 0.5646 \pinkup{\phantom{0}50.12}
        & 16.70
        &  \\
        & SwinIR
        & 0.2466 \pinkdown{\phantom{0}46.07}
        & 0.2424 \pinkdown{\phantom{0}46.29}
        & 22.42 \pinkup{\phantom{0}30.87}
        & 0.5827 \pinkup{\phantom{0}45.46}
        & \underline{0.86}
        &  \\
        & ESRGAN
        & 0.2666 \pinkdown{\phantom{0}50.11}
        & 0.2666 \pinkdown{\phantom{0}51.16}
        & 23.66 \pinkup{\phantom{0}24.01}
        & 0.6426 \pinkup{\phantom{0}31.90}
        & 16.70
        &  \\
        & SinSR
        & 0.2452 \pinkdown{\phantom{0}45.76}
        & 0.2444 \pinkdown{\phantom{0}46.73}
        & 25.44 \pinkup{\phantom{0}15.33}
        & 0.6817 \pinkup{\phantom{0}24.34}
        & 118.51
        &  \\
        & STAR-RL
        & \underline{0.2158} \pinkdown{\phantom{0}38.37}
        & \underline{0.2154} \pinkdown{\phantom{0}39.55}
        & 24.21 \pinkup{\phantom{0}21.19}
        & 0.6784 \pinkup{\phantom{0}24.94}
        & 6.80
        &  \\
        & UPSR
        & 0.3121 \pinkdown{\phantom{0}57.39}
        & 0.3011 \pinkdown{\phantom{0}56.76}
        & 18.65 \pinkup{\phantom{0}57.33}
        & 0.5454 \pinkup{\phantom{0}55.41}
        & 119.34
        &  \\
        & SuperDiff
        & 0.2624 \pinkdown{\phantom{0}49.31}
        & 0.2532 \pinkdown{\phantom{0}48.58}
        & 23.73 \pinkup{\phantom{0}23.65}
        & 0.6527 \pinkup{\phantom{0}29.86}
        & 181.21
        &  \\
        & LIIF
        & 0.2564 \pinkdown{\phantom{0}48.13}
        & 0.2564 \pinkdown{\phantom{0}49.22}
        & \underline{27.41} \pinkup{\phantom{00}7.05}
        & \underline{0.8057} \pinkup{\phantom{00}5.20}
        & 1.57
        &  \\
        & \cellcolor{tablegray}\textbf{Morph-ISR (ours)}
        & \multicolumn{1}{c}{\cellcolor{tablegray}\textbf{0.1330}}
        & \multicolumn{1}{c}{\cellcolor{tablegray}\textbf{0.1302}}
        & \multicolumn{1}{c}{\cellcolor{tablegray}\textbf{29.34}}
        & \multicolumn{1}{c}{\cellcolor{tablegray}\textbf{0.8476}}
        & \multicolumn{1}{r}{\cellcolor{tablegray}\textbf{0.8291}}
        &  \\
        \midrule[0.5mm]

        \multirow{10}{*}{\rotatebox[origin=c]{90}{\textbf{TCGA-KIRC}}}
        & Bicubic
        & 0.3521 \pinkdown{\phantom{0}58.76}
        & 0.3386 \pinkdown{\phantom{0}57.91}
        & \underline{26.23} \pinkup{\phantom{0}13.11}
        & 0.7021 \pinkup{\phantom{0}17.52}
        & --
        & \multirow{10}{*}{\rotatebox[origin=c]{270}{\textbf{FLOPs (G)$\downarrow$}}} \\
        & SHISRCNet
        & 0.2525 \pinkdown{\phantom{0}42.50}
        & 0.2504 \pinkdown{\phantom{0}43.09}
        & 23.04 \pinkup{\phantom{0}28.77}
        & 0.5645 \pinkup{\phantom{0}46.16}
        & 378.18
        &  \\
        & SwinIR
        & 0.2211 \pinkdown{\phantom{0}34.33}
        & 0.2204 \pinkdown{\phantom{0}35.34}
        & 24.06 \pinkup{\phantom{0}23.31}
        & 0.6941 \pinkup{\phantom{0}18.87}
        & \underline{195.60}
        &  \\
        & ESRGAN
        & 0.2121 \pinkdown{\phantom{0}31.54}
        & 0.2121 \pinkdown{\phantom{0}32.81}
        & 22.86 \pinkup{\phantom{0}29.78}
        & 0.6253 \pinkup{\phantom{0}31.95}
        & 293.71
        &  \\
        & SinSR
        & 0.2564 \pinkdown{\phantom{0}43.37}
        & 0.2552 \pinkdown{\phantom{0}44.16}
        & 25.15 \pinkup{\phantom{0}17.97}
        & 0.6614 \pinkup{\phantom{0}24.75}
        & 200.56
        &  \\
        & STAR-RL
        & 0.2361 \pinkdown{\phantom{0}38.50}
        & 0.2358 \pinkdown{\phantom{0}39.57}
        & 26.12 \pinkup{\phantom{0}13.58}
        & 0.6823 \pinkup{\phantom{0}20.93}
        & 445.58
        &  \\
        & UPSR
        & 0.2848 \pinkdown{\phantom{0}49.02}
        & 0.2848 \pinkdown{\phantom{0}49.96}
        & 19.55 \pinkup{\phantom{0}51.76}
        & 0.5555 \pinkup{\phantom{0}48.53}
        & 294.65
        &  \\
        & SuperDiff
        & \underline{0.2104} \pinkdown{\phantom{0}30.99}
        & \underline{0.2104} \pinkdown{\phantom{0}32.27}
        & 23.62 \pinkup{\phantom{0}25.61}
        & 0.6747 \pinkup{\phantom{0}22.29}
        & 503.24
        &  \\
        & LIIF
        & 0.2191 \pinkdown{\phantom{0}33.73}
        & 0.2191 \pinkdown{\phantom{0}34.96}
        & 25.55 \pinkup{\phantom{0}16.12}
        & \underline{0.8019} \pinkup{\phantom{00}2.89}
        & 382.62
        &  \\
        & \cellcolor{tablegray}\textbf{Morph-ISR (ours)}
        & \multicolumn{1}{c}{\cellcolor{tablegray}\textbf{0.1452}}
        & \multicolumn{1}{c}{\cellcolor{tablegray}\textbf{0.1425}}
        & \multicolumn{1}{c}{\cellcolor{tablegray}\textbf{29.67}}
        & \multicolumn{1}{c}{\cellcolor{tablegray}\textbf{0.8251}}
        & \multicolumn{1}{r}{\cellcolor{tablegray}\textbf{189.2}}
        &  \\
        \midrule[0.5mm]

        \multirow{10}{*}{\rotatebox[origin=c]{90}{\textbf{TCGA-LIHC}}}
        & Bicubic
        & 0.3523 \pinkdown{\phantom{0}58.47}
        & 0.3384 \pinkdown{\phantom{0}57.12}
        & 26.23 \pinkup{\phantom{00}7.05}
        & 0.7023 \pinkup{\phantom{0}14.17}
        & --
        & \multirow{10}{*}{\rotatebox[origin=c]{270}{\textbf{Throughput (FPS)$\uparrow$}}} \\
        & SHISRCNet
        & 0.2321 \pinkdown{\phantom{0}36.97}
        & 0.2251 \pinkdown{\phantom{0}35.54}
        & 23.24 \pinkup{\phantom{0}20.82}
        & 0.6542 \pinkup{\phantom{0}22.56}
        & 8.31
        &  \\
        & SwinIR
        & 0.2202 \pinkdown{\phantom{0}33.56}
        & 0.2202 \pinkdown{\phantom{0}34.11}
        & 23.12 \pinkup{\phantom{0}21.45}
        & 0.6646 \pinkup{\phantom{0}20.64}
        & 2.57
        &  \\
        & ESRGAN
        & 0.2332 \pinkdown{\phantom{0}37.26}
        & 0.2332 \pinkdown{\phantom{0}37.78}
        & 23.12 \pinkup{\phantom{0}21.45}
        & 0.6324 \pinkup{\phantom{0}26.79}
        & 8.31
        &  \\
        & SinSR
        & 0.2664 \pinkdown{\phantom{0}45.08}
        & 0.2652 \pinkdown{\phantom{0}45.29}
        & 25.13 \pinkup{\phantom{0}11.73}
        & 0.6514 \pinkup{\phantom{0}23.09}
        & 2.57
        &  \\
        & STAR-RL
        & 0.2531 \pinkdown{\phantom{0}42.20}
        & 0.2531 \pinkdown{\phantom{0}42.67}
        & 24.24 \pinkup{\phantom{0}15.83}
        & 0.6628 \pinkup{\phantom{0}20.97}
        & 4.25
        &  \\
        & UPSR
        & 0.2788 \pinkdown{\phantom{0}47.53}
        & 0.2727 \pinkdown{\phantom{0}46.79}
        & 19.34 \pinkup{\phantom{0}45.18}
        & 0.5155 \pinkup{\phantom{0}55.54}
        & 8.27
        &  \\
        & SuperDiff
        & 0.2492 \pinkdown{\phantom{0}41.29}
        & 0.2492 \pinkdown{\phantom{0}41.77}
        & 23.62 \pinkup{\phantom{0}18.88}
        & 0.6548 \pinkup{\phantom{0}22.45}
        & 0.08
        &  \\
        & LIIF
        & \underline{0.1880} \pinkdown{\phantom{0}22.18}
        & \underline{0.1880} \pinkdown{\phantom{0}22.82}
        & \underline{26.99} \pinkup{\phantom{00}4.03}
        & \textbf{0.8288} \greendown{\phantom{00}3.26}
        & \underline{10.87}
        &  \\
        & \cellcolor{tablegray}\textbf{Morph-ISR (ours)}
        & \multicolumn{1}{c}{\cellcolor{tablegray}\textbf{0.1463}}
        & \multicolumn{1}{c}{\cellcolor{tablegray}\textbf{0.1451}}
        & \multicolumn{1}{c}{\cellcolor{tablegray}\textbf{28.08}}
        & \multicolumn{1}{c}{\cellcolor{tablegray}\underline{0.8018}}
        & \multicolumn{1}{r}{\cellcolor{tablegray}\textbf{11.89}}
        &  \\
        \midrule[0.5mm]

        \multirow{10}{*}{\rotatebox[origin=c]{90}{\textbf{SurGen}}}
        & Bicubic
        & 0.2425 \pinkdown{\phantom{0}28.91}
        & 0.2392 \pinkdown{\phantom{0}28.05}
        & \textbf{38.61} \greendown{\phantom{00}5.05}
        & \textbf{0.9248} \greendown{\phantom{00}3.17}
        & --
        & \multirow{10}{*}{\rotatebox[origin=c]{270}{\textbf{Inference Jitter (ms)$\downarrow$}}} \\
        & SHISRCNet
        & 0.2185 \pinkdown{\phantom{0}21.10}
        & 0.2116 \pinkdown{\phantom{0}18.67}
        & 27.51 \pinkup{\phantom{0}33.26}
        & 0.6925 \pinkup{\phantom{0}29.31}
        & \textbf{0.35}
        &  \\
        & SwinIR
        & 0.1781 \pinkdown{\phantom{00}3.20}
        & 0.1764 \pinkdown{\phantom{00}2.44}
        & 30.15 \pinkup{\phantom{0}21.59}
        & 0.8224 \pinkup{\phantom{00}8.89}
        & 346.66
        &  \\
        & ESRGAN
        & 0.1784 \pinkdown{\phantom{00}3.36}
        & 0.1776 \pinkdown{\phantom{00}3.10}
        & 29.17 \pinkup{\phantom{0}25.68}
        & 0.7366 \pinkup{\phantom{0}21.57}
        & \underline{2.18}
        &  \\
        & SinSR
        & 0.2324 \pinkdown{\phantom{0}25.82}
        & 0.2314 \pinkdown{\phantom{0}25.63}
        & 25.27 \pinkup{\phantom{0}45.07}
        & 0.5586 \pinkup{\phantom{0}60.31}
        & 467.83
        &  \\
        & STAR-RL
        & 0.1842 \pinkdown{\phantom{00}6.41}
        & 0.1842 \pinkdown{\phantom{00}6.57}
        & 30.24 \pinkup{\phantom{0}21.23}
        & 0.7566 \pinkup{\phantom{0}18.36}
        & 5.67
        &  \\
        & UPSR
        & \underline{0.1735} \pinkdown{\phantom{00}0.63}
        & \underline{0.1723} \pinkdown{\phantom{00}0.12}
        & 28.62 \pinkup{\phantom{0}28.09}
        & 0.8125 \pinkup{\phantom{0}10.22}
        & 1775.01
        &  \\
        & SuperDiff
        & 0.1784 \pinkdown{\phantom{00}3.36}
        & 0.1782 \pinkdown{\phantom{00}3.42}
        & 26.58 \pinkup{\phantom{0}37.92}
        & 0.7818 \pinkup{\phantom{0}14.54}
        & 1556.31
        &  \\
        & LIIF
        & 0.2226 \pinkdown{\phantom{0}22.55}
        & 0.2161 \pinkdown{\phantom{0}20.36}
        & 35.99 \pinkup{\phantom{00}1.86}
        & 0.8920 \pinkup{\phantom{00}0.39}
        & 4.42
        &  \\
        & \cellcolor{tablegray}\textbf{Morph-ISR (ours)}
        & \multicolumn{1}{c}{\cellcolor{tablegray}\textbf{0.1724}}
        & \multicolumn{1}{c}{\cellcolor{tablegray}\textbf{0.1721}}
        & \multicolumn{1}{c}{\cellcolor{tablegray}\underline{36.66}}
        & \multicolumn{1}{c}{\cellcolor{tablegray}\underline{0.8955}}
        & \multicolumn{1}{r}{\cellcolor{tablegray}7.153}
        &  \\

        \bottomrule[0.8mm]
    \end{tabular}%
    \caption{Super-resolution performance of various frameworks on the TCGA and SurGen datasets. Our proposed method is shaded in \colorbox[HTML]{EFEFEF}{gray}. For each compared method, the percentage corresponding to {\color[HTML]{BE129D}↓} or {\color[HTML]{036400}↑} indicates the relative difference in performance compared to ours. Additionally, the best-performing results are highlighted in \textbf{bold}, whereas the second-best results are \uline{underlined}.}
    \label{tab:benchmark}
\end{table*}

\begin{figure*}[!t]
    \centering      
    \includegraphics[width=\textwidth]{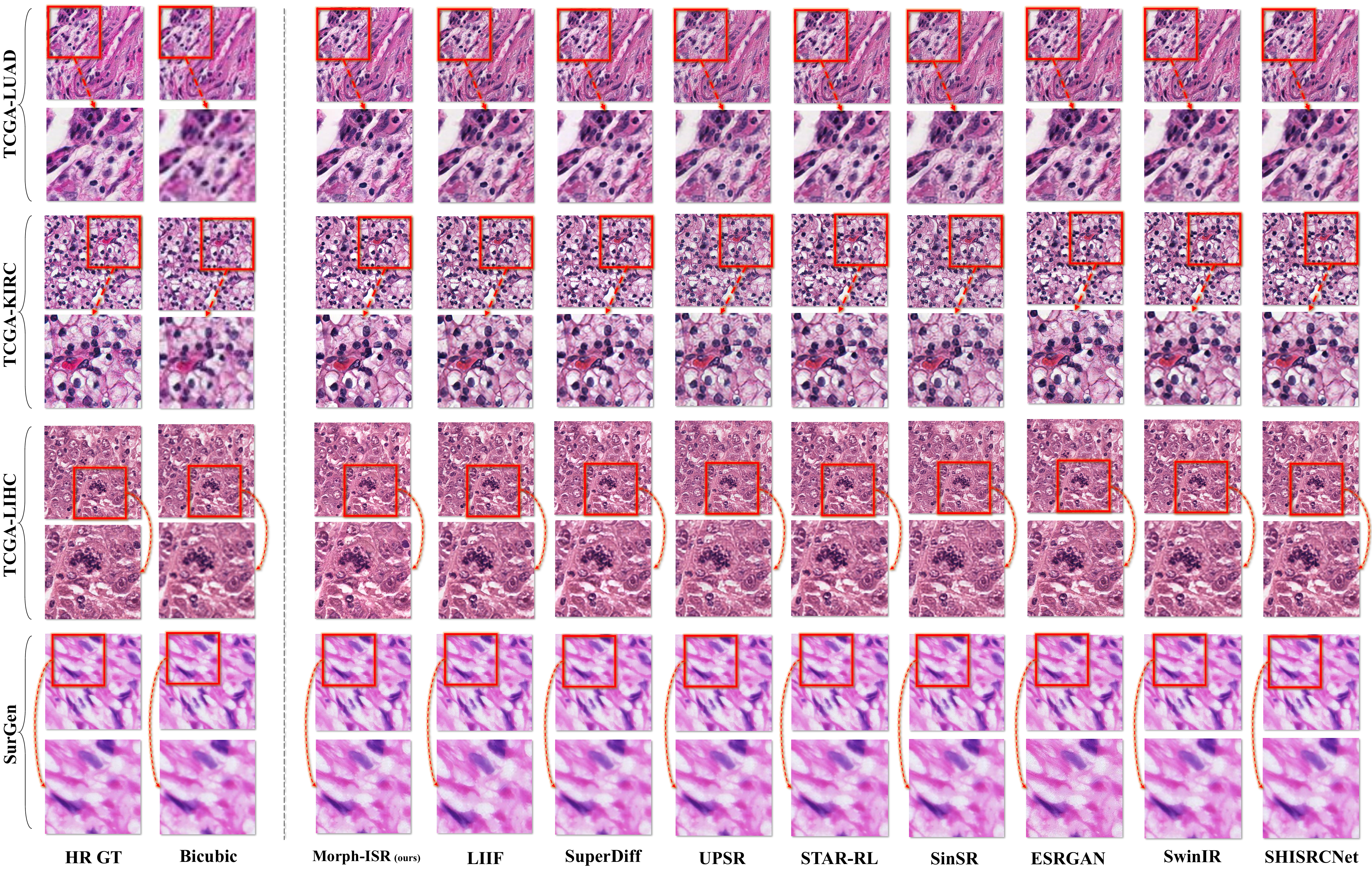}
    \caption{Qualitative visualizations of super-resolution in pathological images.}
    \label{fig:QV}
\end{figure*}

\subsection{Implicit Position-aware Kernel Generator Module}
To address the geometric discontinuity challenge inherent in cross-scale reconstruction, the IPKG module is established as a gated multi-head kernel generator that dynamically synthesizes spatially adaptive resampling kernels for each continuous query coordinate. For the $i$-th query point, IPKG takes three position-aware inputs defined in the previous subsection: the center feature $f_{cen,i}$ sampled from the LR latent feature map at $x_{c,i}$, the Fourier positional feature $x_{pe,i}=\gamma(x_{nc,i})$, and the sub-pixel phase shift $x_{sp,i}$. These features jointly describe local texture, global coordinate position, and sub-pixel alignment. To accommodate the morphological diversity of pathological tissue, ranging from dense chromatin to sparse stroma, a gated multi-head architecture is adopted.

A shared lightweight MLP $\Phi(\cdot)$ maps the composite query feature $v_{q,i}\in\mathbb{R}^{C+P+2}$ into $N_h$ pairs of raw kernel and gate logits, where $N_h$ denotes the number of kernel heads. The kernel logits $\mathcal{W}_{raw,i}^{(h)}\in\mathbb{R}^{K^2}$ parameterize the $K\times K$ filter of the $h$-th head, while $\mathcal{G}_{raw,i}^{(h)}\in\mathbb{R}$ represents its confidence for the current query. Their computation is given by Eq. (\ref{eq: wg1})-(\ref{eq: wg2}):
\begin{equation}
\begin{aligned}
\label{eq: wg1}
    &v_{q,i} = f_{cen,i}\oplus x_{pe,i}\oplus x_{sp,i}.
\end{aligned}
\end{equation}
\begin{equation}
\begin{aligned}
\label{eq: wg2}
    \{ \mathcal{W}_{raw,i}^{(h)}, \mathcal{G}_{raw,i}^{(h)} \}_{h=1}^{N_h} = \Phi(v_{q,i}).
\end{aligned}
\end{equation}

To reduce ambiguity in head selection while ensuring that the learned weights form valid convolutional filters, a temperature-scaled normalization scheme is introduced. Specifically, the gate logits are normalized over the $N_h$ heads to produce $\alpha_{gate,i}^{(h)}$, whereas the kernel logits of each head are normalized over the $K^2$ spatial positions to obtain $w_{head,i}^{(h)}$. The two normalization operations are controlled by temperatures $\tau_1$ and $\tau_2$, and are formulated as Eq. (\ref{eq: moe_norm1})-(\ref{eq: moe_norm2}):
\begin{equation}
\label{eq: moe_norm1}
\begin{aligned}
    \alpha_{gate,i}^{(h)}
    &=
    \frac{
    \exp\!\left(\mathcal{G}_{raw,i}^{(h)}/\tau_1\right)}
    {
    \sum_{j=1}^{N_h}
    \exp\!\left(\mathcal{G}_{raw,i}^{(j)}/\tau_1\right)
    }.
\end{aligned}
\end{equation}

\begin{equation}
\label{eq: moe_norm2}
\begin{aligned}
    w_{head,i}^{(h)}
    &=
    \operatorname{Softmax}\!\left(
    \frac{\mathcal{W}_{raw,i}^{(h)}}{\tau_2}
    \right).
\end{aligned}
\end{equation}

Finally, the spatially adaptive kernel $w_i\in\mathbb{R}^{K^2}$ is derived through confidence-weighted head aggregation. This dynamically synthesized kernel is subsequently applied to the local neighborhood vector $V_i$ defined in Eq. (\ref{eq: reconstruction1}), enabling query-specific restoration of heterogeneous pathological textures while maintaining sub-pixel geometric continuity. The aggregation is expressed in Eq. (\ref{eq: wi}):
\begin{equation}
\label{eq: wi}
w_i=
\sum_{h=1}^{N_h}
\alpha_{gate,i}^{(h)}
w_{head,i}^{(h)}.
\end{equation}
Collecting all query kernels yields $\Omega=[w_1^\top,\ldots,w_N^\top]^\top\in\mathbb{R}^{N\times K^2}$, which denotes the dynamic kernel matrix shown in Figure.~\ref{fig:overview} (b).

\subsection{Morphological Fidelity Prior Mechanism}
The MFP mechanism constrains the reconstruction $\hat I_{sr}$ using the HR target $I_{hr}$ to preserve diagnostically relevant features. Initially, to counteract spectral bias while preserving perceptual texture and local structure, a Pixel-Spectral Fidelity Loss is formulated in Eq. (\ref{eq: pix}). This term integrates the spatial $L_1$ distance for global photometric accuracy, a frequency-domain constraint based on Fast Fourier Transform (FFT), perceptual feature distance~\cite{zhang2018unreasonable}, and structural similarity consistency~\cite{wang2004image}. By default, the spectral balancing coefficient is set to $\lambda_{freq}=0.05$, while $\lambda_{\phi}=0.15$ and $\lambda_{\rho}=1.0$ balance perceptual and structural consistency. The resulting loss is formulated in Eq. (\ref{eq: pix}):
\begin{equation}
\label{eq: pix}
\begin{aligned}
\mathcal{L}_{pix}
=&
\left\|\hat I_{sr}-I_{hr}\right\|_1
+
\lambda_{freq}
\left\|
\mathcal{F}_{\nu}(\hat I_{sr})-\mathcal{F}_{\nu}(I_{hr})
\right\|_1 \\
&+
\lambda_{\phi}\mathcal{D}_{\phi}(\hat I_{sr},I_{hr})
+
\lambda_{\rho}\left(1-\rho(\hat I_{sr},I_{hr})\right).
\end{aligned}
\end{equation}
where $\|\cdot\|_1$ denotes the mean absolute difference, $\mathcal{F}_{\nu}(\cdot)$ denotes grayscale high-frequency FFT magnitude extraction, $\mathcal{D}_{\phi}(\cdot,\cdot)$ denotes perceptual feature distance~\cite{zhang2018unreasonable}, and $\rho(\cdot,\cdot)$ denotes structural similarity~\cite{wang2004image}. Crucially, to ensure the integrity of cellular structures and the sharpness of geometric contours, semantic priors derived from the HoVer-Net \cite{graham2019hover} are integrated into the optimization process. Specifically, the nuclear mask map $I_{m}$ indicating nuclear interiors and the boundary map $I_{b}$ delineating nuclear contours are extracted to serve as spatial attention weights. These maps guide the model to prioritize the restoration of chromatin textures within cell bodies and structural transitions at high-frequency boundaries, respectively. The corresponding Region-Masked Loss and Boundary-Structural Loss are shown in Eq. (\ref{eq: hov1})-(\ref{eq: hov2}):
\begin{equation}
\label{eq: hov1}
\begin{aligned}
\mathcal{L}_{mask}
&=
\frac{
\sum\left(
I_m\odot\left|\hat I_{sr}-I_{hr}\right|
\right)
}{
\sum I_m+\epsilon
}.
\end{aligned}
\end{equation}
\begin{equation}
\label{eq: hov2}
\begin{aligned}
\mathcal{L}_{bound}
=
\frac{
\sum\left(
I_b\odot
\left|
\mathcal{H}(\hat I_{sr})-\mathcal{H}(I_{hr})
\right|
\right)
}{
\sum I_b+\epsilon
}.
\end{aligned}
\end{equation}
where $\odot$ denotes the Hadamard product, and $\mathcal{H}(\cdot)$ represents a grayscale high-pass operator for extracting local edge responses. Finally, the total objective is defined as the weighted sum: $\mathcal{L}_{total}=\mathcal{L}_{pix}+\lambda_{mask}\mathcal{L}_{mask}+\lambda_{bound}\mathcal{L}_{bound}$. In our experiments, the weighting coefficients for region and boundary constraints are set to $\lambda_{mask}=0.65$ and $\lambda_{bound}=1.0$ by default, emphasizing morphological fidelity.

\section{Experiments}

\subsection{Experiment Settings}

\textbf{Datasets and Implementation: }
The experiments use 150 H\&E-stained TCGA WSIs, comprising 50 randomly selected cases from each of the LUAD, KIRC, and LIHC cohorts~\cite{weinstein2013cancer}. These WSIs are divided into mutually exclusive training, validation, and test sets using a 7:1:2 ratio. CLAM~\cite{lu2021data} further extracts 34,495 HR patches of $512\times512$ pixels at $40\times$ magnification, which are bicubically downsampled by $4\times$ to produce paired $128\times128$ LR inputs. External generalization is assessed by directly applying the TCGA-trained model to the independent SurGen cohort of 46 colorectal cancer patients~\cite{myles2025surgen}, which is excluded from training and model selection but undergoes the same preprocessing.

Under this experimental protocol, the framework is implemented in PyTorch and trained on a single NVIDIA RTX PRO 6000 GPU for 120 epochs. Optimization uses AdamW with a batch size of 4, a learning rate of $3\times10^{-4}$, and a weight decay of $10^{-4}$. Automatic mixed precision and gradient clipping at 1.0 are applied throughout training. Meanwhile, the gating and kernel temperatures $(\tau_1,\tau_2)$ are progressively annealed from $(4.0,1.0)$ to $(1.2,0.6)$ to stabilize dynamic kernel learning. Further details regarding the datasets and implementation are provided in Supplementary Appendix A and B.

\noindent

\noindent

\textbf{Competing Methods and Metrics:}
To comprehensively evaluate Morph-ISR, the baselines comprise \textbf{Bicubic} interpolation, deterministic regression methods SHISRCNet~\cite{xie2023shisrcnet} and SwinIR~\cite{liang2021swinir}, generative reconstruction methods ESRGAN~\cite{wang2018esrgan}, SinSR~\cite{wang2024sinsr}, STAR-RL~\cite{chen2024star}, UPSR~\cite{zhang2025uncertainty}, and SuperDiff~\cite{xu2025superdiff}, and the continuous SR method LIIF~\cite{chen2021learning}. For quantitative validation,  LPIPS and ST-LPIPS~\cite{zhang2018perceptual} assess perceptual and structural consistency, while PSNR and SSIM~\cite{wang2004image} measure full-reference fidelity. Complementing the accuracy evaluation, edge-oriented deployment efficiency is characterized by parameters, FLOPs, throughput, and inference jitter following~\cite{pang2025slim}.

To determine whether the SR preserves pathology-relevant structures, downstream nuclear analysis is further conducted on representative images from TCGA-LUAD, TCGA-KIRC, TCGA-LIHC, and SurGen. A fixed H\&E Cellpose cyto3 segmentation pipeline~\cite{stringer2025cellpose3}, independent of the HoVer-Net supervision employed in MFP, generates nuclear instances from SR and the corresponding HR images. Using the HR-derived segmentations as references, Nuclear Mask Dice and Detection F1 evaluate segmentation consistency, while Circularity and Eccentricity Fidelity quantify morphological preservation. This inference-only analysis requires no additional training or parameter updates. Further dataset, implementation, and computational details are provided in Supplementary Appendix C.

\subsection{Results and Discussion}
\textbf{Quantitative Analysis of SR}: 
Table~\ref{tab:benchmark} shows that Morph-ISR achieves the best LPIPS and ST-LPIPS across all three TCGA cohorts, with the largest margins on TCGA-LUAD: reductions of 38.37\% and 39.55\% relative to the second-best method, STAR-RL. These gains hold across deterministic and generative baselines as well as the continuous SR method LIIF, indicating robustness across reconstruction formulations. Figure~\ref{fig:QV} shows the same trend. Bicubic interpolation and deterministic baselines blur nuclear boundaries and attenuate stromal textures, while several generative methods produce less stable local morphology. In contrast, Morph-ISR better preserves nuclear contours, chromatin patterns, and fine tissue structures without conspicuous artificial textures. The independent SurGen evaluation further supports external-domain generalization. Morph-ISR achieves the best LPIPS and ST-LPIPS scores, at 0.1724 and 0.1721, respectively, improving over the second-best method, UPSR, by 0.63\% and 0.12\%, respectively. It also achieves the best PSNR across the three TCGA cohorts, the best SSIM on TCGA-LUAD and TCGA-KIRC, and the second-best SSIM on TCGA-LIHC, while ranking second in both PSNR and SSIM on SurGen. Although Bicubic leads these two pixel-based metrics on SurGen, its substantially poorer perceptual scores reveal the limitations of pixel fidelity alone. Morph-ISR further uses 0.8291 M parameters and 189.2 G total FLOPs, achieves the highest throughput of 11.89 FPS, and records 7.153 ms inference jitter, supporting edge-oriented deployment. Overall, Morph-ISR combines perceptual and structural fidelity with external generalization and deployment efficiency.

\begin{figure}[!t]
    \centering      
    \includegraphics[width=\columnwidth]{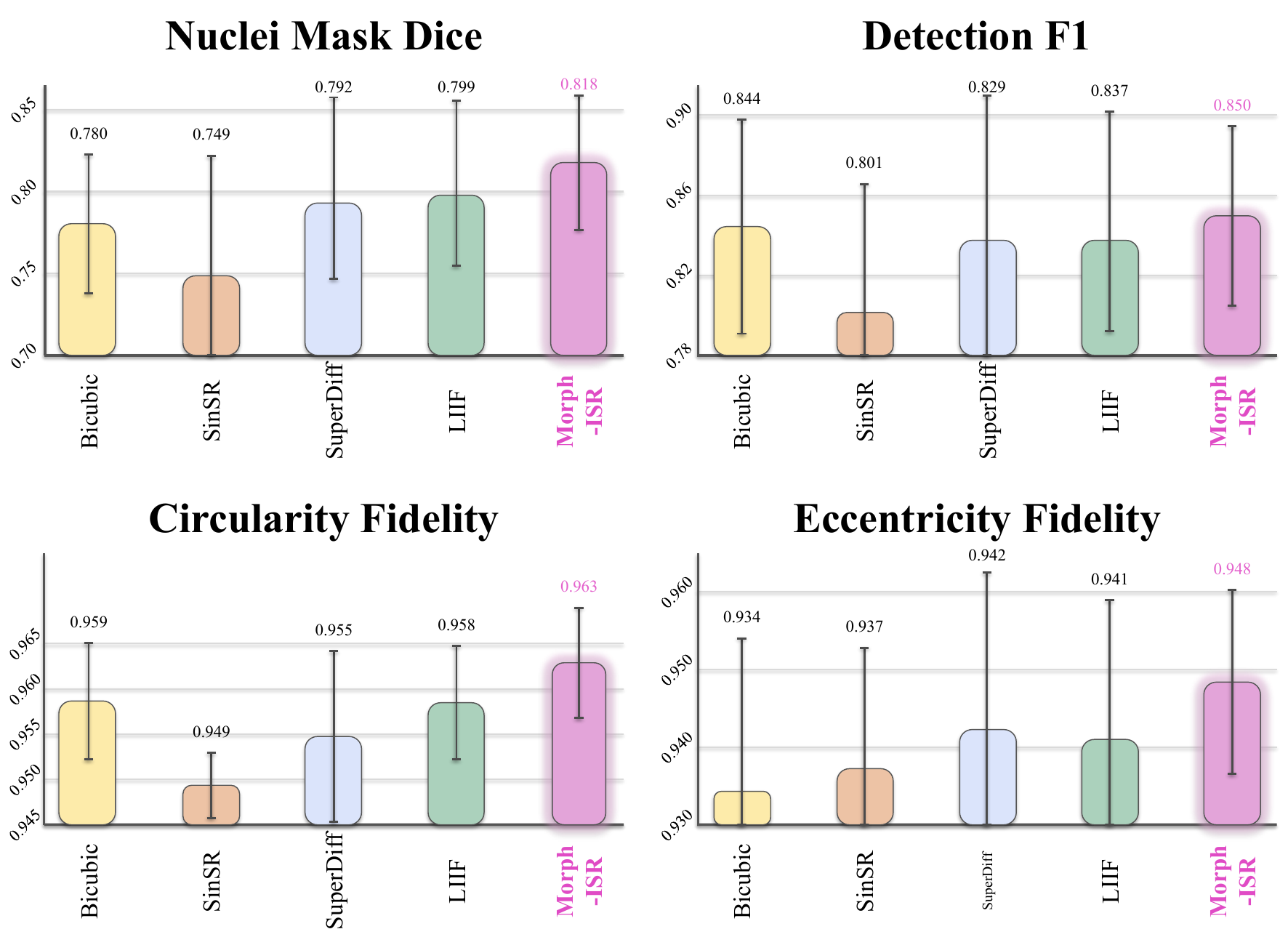}
    \caption{Quantitative comparison of downstream nuclei segmentation consistency and morphological fidelity.}
    \label{fig:nmb}
\end{figure}

\begin{figure}[!t]
    \centering      
    \includegraphics[width=\columnwidth]{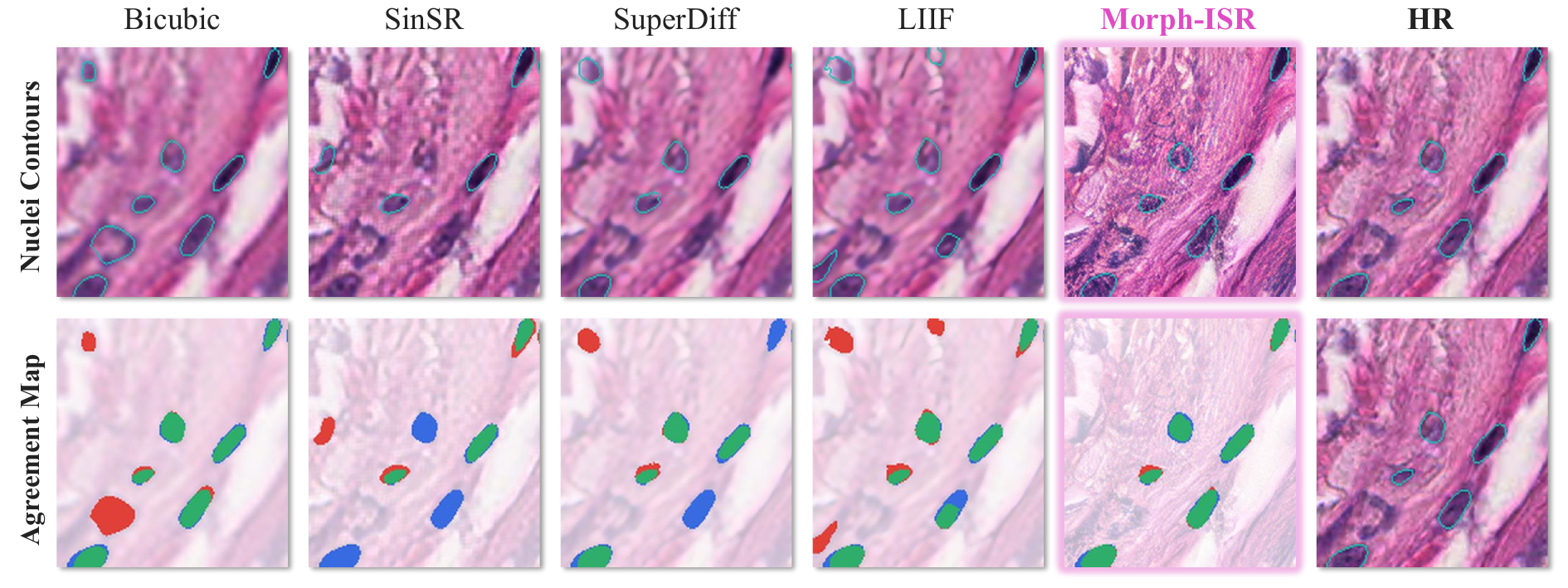}
    \caption{Visual comparison of downstream nuclei analysis on a representative TCGA-LUAD region. Cyan contours mark detected nuclear boundaries. In the agreement maps, green, red, and blue denote matched nuclei regions, false-positive regions, and false-negative regions, respectively.}
    \label{fig:nv}
\end{figure}

\textbf{Downstream Nuclei Analysis}:
As shown in Fig.~\ref{fig:nmb}, Morph-ISR records the highest mean scores among the evaluated methods across all four downstream metrics in nuclei analysis. It achieves a nuclear mask Dice of 0.818 and a Detection F1 of 0.850, indicating the closest segmentation agreement with the HR-derived Cellpose references. Its Circularity Fidelity and Eccentricity Fidelity reach 0.963 and 0.948, respectively, indicating more faithful preservation of nuclear shape characteristics. The consistent mean-score advantage across segmentation and morphology metrics supports the effectiveness of Morph-ISR in retaining pathology-relevant nuclear structures. This quantitative trend is further supported by Figure.~\ref{fig:nv}, where Morph-ISR produces nuclear contours closely aligned with the HR reference and coherent agreement regions with fewer localized mismatches. Together, these results indicate that Morph-ISR preserves diagnostically relevant nuclear morphology beyond merely producing sharper visual appearances.

\begin{table}[!t]
    \centering
    \renewcommand{\arraystretch}{1.08}
    \setlength{\tabcolsep}{3.6pt}
    \begin{tabular}{l|ccc}
        \toprule[0.8mm]
        \multicolumn{1}{c|}{\multirow{2}{*}{\textbf{Metric}}} &
        \multicolumn{3}{c}{\textbf{Training setting}} \\
        \cmidrule(lr){2-4}
        & $(\checkmark,\times,\times)$
        & $(\checkmark,\checkmark,\times)$
        & \cellcolor{tablegray}$(\checkmark,\checkmark,\checkmark)$ \\
        \midrule[0.5mm]

        LPIPS $\downarrow$
        & 0.3430 \pinkdown{61.2}
        & \underline{0.1887} \pinkdown{29.5}        & \cellcolor{tablegray}\textbf{0.1330} \\

        ST-LPIPS $\downarrow$
        & 0.3420 \pinkdown{61.9}
        & \underline{0.1880} \pinkdown{30.7}
        & \cellcolor{tablegray}\textbf{0.1302} \\

        \addlinespace[1pt]
        PSNR $\uparrow$
        & 25.84\phantom{0} \pinkup{13.6}
        & \underline{27.56}\phantom{0} \pinkup{\phantom{0}6.5}
        & \cellcolor{tablegray}\textbf{29.34} \\

        SSIM $\uparrow$
        & 0.6947 \pinkup{22.0}
        & \underline{0.8042} \pinkup{\phantom{0}5.4}
        & \cellcolor{tablegray}\textbf{0.8476} \\

        \bottomrule[0.8mm]
    \end{tabular}%
    \caption{Ablation study of Morph-ISR on TCGA-LUAD. Columns denote $(\mathcal{L}_{pix},\mathrm{IPKG},\mathrm{MFP})$. The highlighted settings in this table are the same as in Table~\ref{tab:benchmark}.}
    \label{tab:AS}
\end{table}

\subsection{Ablation Study}
Table~\ref{tab:AS} demonstrates the complementary contributions of IPKG and MFP. Adding IPKG to the pixel-loss baseline reduces LPIPS from 0.3430 to 0.1887 and ST-LPIPS from 0.3420 to 0.1880, while increasing SSIM from 0.6947 to 0.8042. These improvements show that position-aware kernels enhance perceptual and structural fidelity. MFP further lowers LPIPS and ST-LPIPS to 0.1330 and 0.1302, corresponding to reductions of 29.52\% and 30.74\% over the IPKG variant. It also increases PSNR from 27.56 to 29.34 dB and SSIM to 0.8476, confirming that morphology-aware supervision strengthens structural preservation while retaining strong pixel-level fidelity. Overall, the full model reduces LPIPS and ST-LPIPS by 61.22\% and 61.93\% over the baseline while improving PSNR and SSIM by 13.55\% and 22.01\%. These progressive gains confirm that IPKG enables spatially adaptive reconstruction across heterogeneous tissue regions, while MFP improves the preservation of diagnostically relevant fine-grained morphology.

\section{Conclusion}
This work presents Morph-ISR, a morphology-aware implicit framework balancing pixel-level accuracy and structural fidelity in pathological image SR. In this framework, IPKG formulates reconstruction as continuous coordinate querying and generates spatially adaptive kernels for heterogeneous tissue structures, while the training-only MFP imposes region- and boundary-aware constraints without inference overhead. Evaluations on three TCGA cohorts show leading perceptual and structural fidelity alongside strong pixel-level accuracy. Zero-shot results on SurGen support cross-dataset generalization, while downstream nuclear analysis supports improved segmentation consistency and morphological preservation. Compact parameterization and favorable throughput further support efficient deployment in resource-constrained settings.

%

%
%
%

\clearpage
\appendix
\section{Supplementary Appendix}
\section{A. Implementation and Reproducibility}
\subsection{A.1 Optimization Details}
Morph-ISR is trained on the TCGA cohorts using the patient-level training split described below. The image reconstruction, perceptual, and morphology-aware losses are jointly optimized with the relative weights specified in the main paper. We use AdamW with the learning-rate schedule, batch size, crop resolution, and total training iterations reported in the main manuscript. All model-selection decisions are made on the held-out validation patients only; the test patients are not used for checkpoint selection or hyperparameter tuning.

\subsection{A.2 Inference Protocol}
At inference, a low-resolution input patch is processed by the trained Morph-ISR model to produce its super-resolved counterpart. Unless otherwise noted, all compared methods use the same input and output resolutions, the same patch-extraction procedure, and the same evaluation masks. Deployment efficiency is reported through parameter count, FLOPs, throughput, and inference jitter, all measured using the same input-resolution setting across methods.

\section{B. Data Preparation and Validation Protocol}
\subsection{B.1 TCGA Cohorts and Patient-level Splits}
We use H\&E-stained whole-slide images from TCGA-LUAD, TCGA-KIRC, and TCGA-LIHC \cite{weinstein2013cancer}. Splits are made at the patient level to prevent patches from the same patient appearing in more than one split. The TCGA data are used for training, validation, and in-domain testing only.

\subsection{B.2 Patch Extraction and Resolution Degradation}
Tissue regions are identified before patch extraction. High-resolution patches are sampled from tissue-containing regions and degraded by the same $4\times$ procedure used for all methods in the main paper. The resulting low-resolution patches are paired with their high-resolution counterparts. No test-set images are used to fit stain normalization, degradation parameters, or any other preprocessing component.

\subsection{B.3 Independent External Evaluation}
\begin{figure*}[t]
\centering
\includegraphics[width=\textwidth]{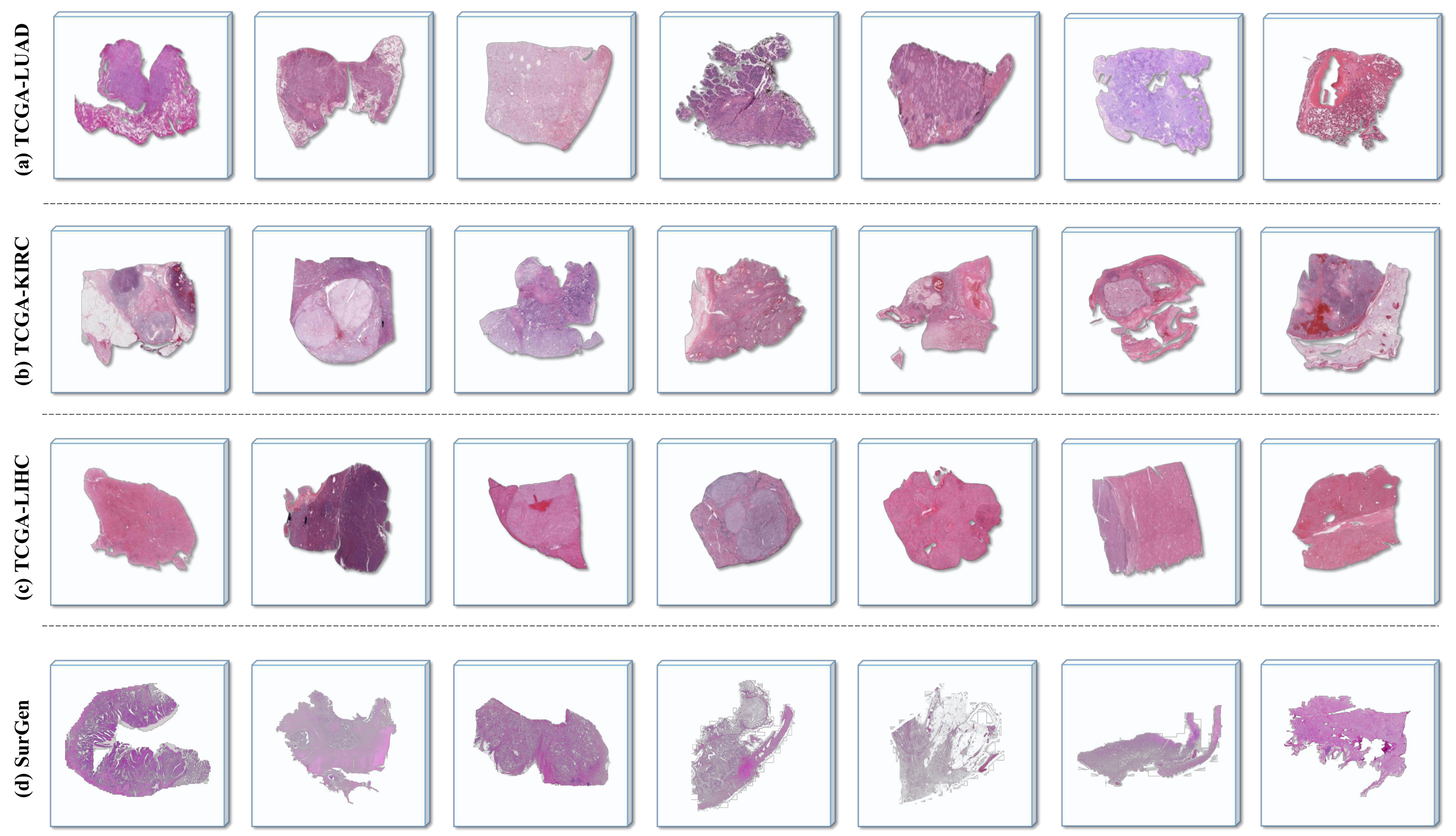}
\caption{Representative H\&E-stained whole-slide images from the TCGA-LUAD, TCGA-KIRC, TCGA-LIHC, and SurGen cohorts, illustrating inter-cohort variation in tissue morphology and staining.}
\label{fig:cohort-overview}
\end{figure*}

Cross-dataset generalization is evaluated on the independent SurGen colorectal-cancer cohort \cite{myles2025surgen}, which contains 46 patients in the prepared evaluation set. The TCGA-trained checkpoint is frozen: SurGen contributes neither to training, validation-based checkpoint selection, nor hyperparameter tuning. The same $4\times$ degradation and patch-extraction procedure is applied. To make external evaluation reproducible while limiting slide imbalance, at most 300 paired patches per slide are retained by deterministic stable-hash sampling with seed 2025. HR images are used only as evaluation references. Representative whole-slide images from the in-domain and external cohorts are shown in Figure.~\ref{fig:cohort-overview}.

\section{C. Evaluation Protocol}
\subsection{C.1 Image Fidelity and Efficiency}
Image reconstruction is assessed using LPIPS and shift-tolerant LPIPS (ST-LPIPS) for perceptual similarity \cite{zhang2018unreasonable}, together with PSNR and SSIM for pixel-level and structural fidelity \cite{wang2004image}. Lower LPIPS and ST-LPIPS indicate better perceptual agreement, while higher PSNR and SSIM indicate better pixel-level fidelity.

Let $x$ and $y$ denote an SR image and its HR reference, respectively, each with $N$ pixels and intensity range $[0,L]$. We use
\begin{equation}
\operatorname{MSE}(x,y)=\frac{1}{N}\sum_{i=1}^{N}(x_i-y_i)^2.
\end{equation}
\begin{equation}
\operatorname{PSNR}(x,y)=10\log_{10}\left(\frac{L^2}{\operatorname{MSE}(x,y)}\right).
\end{equation}
Structural similarity is computed as
\begin{equation}
\operatorname{SSIM}(x,y)=\frac{(2\mu_x\mu_y+C_1)(2\sigma_{xy}+C_2)}
{(\mu_x^2+\mu_y^2+C_1)(\sigma_x^2+\sigma_y^2+C_2)},
\end{equation}
where $\mu$, $\sigma^2$, and $\sigma_{xy}$ are local means, variances, and covariance, and $C_1,C_2$ are stabilizing constants. For normalized deep features $\hat f_l$, define
\begin{equation}
\Delta_{l,hw}(x,y)=\hat f_l(x)_{hw}-\hat f_l(y)_{hw}.
\end{equation}
LPIPS then compares the feature differences as
\begin{equation}
\operatorname{LPIPS}(x,y)=\sum_l\frac{1}{H_lW_l}
\sum_{h,w}\left\|w_l\odot\Delta_{l,hw}(x,y)\right\|_2^2,
\end{equation}
where $w_l$ denotes learned channel weights. ST-LPIPS is evaluated by allowing prescribed small spatial translations:
\begin{equation}
\operatorname{ST\text{-}LPIPS}(x,y)=\min_{\delta\in\mathcal{D}}\operatorname{LPIPS}(T_\delta(x),y),
\end{equation}
where $T_\delta$ translates the image by $\delta$ and $\mathcal{D}$ is the evaluator's allowed shift set.

\subsection{C.2 Nuclear Morphology Assessment}
To evaluate whether restored images preserve diagnostically relevant cellular structures, a fixed H\&E Cellpose cyto3 segmentation pipeline~\cite{stringer2025cellpose3} is applied to each SR image and its corresponding HR image to generate nuclear instances. This evaluation pipeline is independent of the HoVer-Net supervision employed by MFP during training~\cite{graham2019hover}. Nuclear Mask Dice and Detection F1 quantify segmentation consistency, while Circularity Fidelity and Eccentricity Fidelity assess morphology preservation. This analysis is inference-only and introduces no additional training.

With predicted and reference nuclear masks $P$ and $G$, Nuclear Mask Dice and Detection F1 are
\begin{equation}
\operatorname{Dice}(P,G)=\frac{2|P\cap G|}{|P|+|G|}.
\end{equation}
\begin{equation}
\operatorname{F1}=\frac{2\operatorname{TP}}{2\operatorname{TP}+\operatorname{FP}+\operatorname{FN}}.
\end{equation}
For matched nuclear instances $j=1,\ldots,M$, circularity is $c_j=4\pi A_j/p_j^2$, where $A_j$ and $p_j$ are area and perimeter. Circularity and eccentricity fidelities are
\begin{equation}
F_c=1-\frac{1}{M}\sum_{j=1}^{M}|c_j^{\mathrm{SR}}-c_j^{\mathrm{HR}}|.
\end{equation}
\begin{equation}
F_e=1-\frac{1}{M}\sum_{j=1}^{M}|e_j^{\mathrm{SR}}-e_j^{\mathrm{HR}}|.
\end{equation}
Here $e_j\in[0,1]$ is the eccentricity of instance $j$; higher Dice, F1, $F_c$, and $F_e$ indicate better preservation.

\section{Scope of the Supplementary Document}
This document supplies reproducibility and evaluation details supporting the main manuscript. The main paper remains self-contained: all information essential to assess the method and its primary claims is included in the main manuscript.

%

\end{document}